\PassOptionsToPackage{dvipsnames}{xcolor}

\documentclass[10pt,technote, compsoc]{IEEEtran}

\ifCLASSOPTIONcompsoc
  \usepackage[nocompress]{cite}
\else
  \usepackage{cite}
\fi

\usepackage{algorithmicx}
\usepackage{graphicx}
\usepackage{textcomp}
\usepackage{graphics}
\usepackage{enumitem}    
\usepackage[ruled,lined]{algorithm2e}
\usepackage{subcaption}
\usepackage{mathtools}
\usepackage{amsmath}
\usepackage{multirow}
\usepackage{array}
\usepackage{siunitx}
\usepackage{booktabs}
\usepackage{threeparttable}
\usepackage{layouts}

\usepackage{svg}

\usepackage{multirow,bigdelim}
\usepackage{rotating}
\usepackage{url}

\usepackage{multirow,bigdelim}
\usepackage{makecell}
\usepackage{rotating}
\newif\ifoutline
\outlinefalse

\newcommand{\blue}[1]{\ifoutline{\color{blue}#1}\fi}

\newcommand{\orange}[1]{{\color{black}#1}}
\newcommand{\green}[1]{{\color{black}#1}}
\newcommand{\yellow}[1]{{\color{black}#1}}

\usepackage{fontawesome}
\usepackage{amsfonts}% http://ctan.org/pkg/amssymb
\usepackage{pifont}% http://ctan.org/pkg/pifont
\newcommand{\cmark}{\color{OliveGreen} \ding{51}}%
\newcommand{\xmark}{\color{BrickRed} \ding{55}}%
\newcommand{\bata}{\color{Orange} \faBattery[1] }%
\newcommand{\batb}{\color{Goldenrod} \faBattery[2]  }%
\newcommand{\batd}{\color{OliveGreen} \faBattery[4]  }%

\usepackage{multirow,bigdelim}
\usepackage{rotating}
\definecolor{Maroon}{cmyk}{0, 0.87, 0.68, 0.32}
\definecolor{RoyalBlue}{cmyk}{1, 0.50, 0, 0}

\begin{document}

\bstctlcite{IEEEexample:BSTcontrol}
\title{Co-Design of Approximate Multilayer Perceptron for Ultra-Resource Constrained Printed Circuits}

\author{Giorgos~Armeniakos,
        Georgios~Zervakis,\\
        Dimitrios~Soudris,~\IEEEmembership{Member,~IEEE,}
        Mehdi~B.~Tahoori,~\IEEEmembership{Fellow,~IEEE,}
        and J{\"o}rg~Henkel,~\IEEEmembership{Fellow,~IEEE}% <-this % stops a space
\IEEEcompsocitemizethanks{\IEEEcompsocthanksitem G.~Armeniakos and D.~Soudris are with the School of Electrical and Computer Engineering, National Technical University of Athens, Athens 15780, Greece. G.~Armeniakos is also with the Chair for Embedded System (CES), Department of Computer Science at Karlsruhe Institute of Technology (KIT), Karlsruhe 76131, Germany.
\IEEEcompsocthanksitem G. Zervakis is with the Computer Engineering \& Informatics Dept., University of Patras, Patras 26504, Greece. This research was done when he was with the Karlsruhe Institute of Technology (KIT).
\IEEEcompsocthanksitem M.~B.~Tahoori is with Chair of Dependable Nano Computing (CDNC), Department of Computer Science at KIT, Karlsruhe 76131, Germany.
\IEEEcompsocthanksitem J.~Henkel is with Chair for Embedded System (CES), Department of Computer Science at KIT, Karlsruhe 76131, Germany.}
\thanks{This work is partially supported by the German Research Foundation through the project “ACCROSS: Approximate Computing aCROss the System Stack” HE 2343/16-1.}%
\thanks{Corresponding Author: Giorgos Armeniakos (georgios.armeniakos@kit.edu)}
\thanks{Manuscript received Mars 24, 2022.}
}

\IEEEtitleabstractindextext{%
\begin{abstract}
Printed Electronics (PE) exhibits on-demand, extremely low-cost hardware due to its additive manufacturing process, enabling machine learning (ML) applications for domains that feature ultra-low cost, conformity, and non-toxicity requirements
that silicon-based systems cannot deliver.
% \orange{
% Such low-cost computing systems pave the way for high customizations (i.e., bespoke circuits), which in silicon-based systems would not be realizable.
% }
Nevertheless, large feature sizes in PE prohibit the realization of complex printed ML circuits.
In this work, we present, for the first time, an automated printed-aware software/hardware co-design framework that exploits approximate computing principles to enable ultra-resource constrained printed multilayer perceptrons (MLPs).
Our evaluation demonstrates that, compared to the state-of-the-art baseline, our circuits feature on average 6x (5.7x) lower area (power) and less than 1\% accuracy loss.

\end{abstract}
}

\maketitle

\IEEEdisplaynontitleabstractindextext
\IEEEpeerreviewmaketitle

\section{Introduction}\label{sec:intro}
% \blue{
% \textbf{Length: 1.25 pages (incl. title+abstract). Figures: 0}
% \begin{itemize}
%     \item Brief introduction for printed electronics machine learning on printed electronics 
%     \item Background on Printed Electronics and Libraries. 
%     \item Highlight area/power problem in printed electronics.
%     \item Brief introduction for Approximate Computing.
%     \item Present our contribution
%     \item Mention that our ML circuit implementations will become available on github
% \end{itemize}
% }
% textwidth in cm: \printinunitsof{cm}\prntlen{\columnwidth}
A trillion-dollar market of fast moving consumer goods (FMCG), disposables such as beverages, packaged foods, and toiletries, low-end healthcare products (e.g., smart bandages), etc., have seen limited impact from embedded computing~\cite{Bleier:ISCA:2020:printedmicro}.
Although for several of these domains, a typically required computational task is classification~\cite{Mubarik:MICRO:2020:printedml}, such applications pose requirements for sub-cent costs, non-toxicity, porosity, stretchability, and conformity that silicon-based computing systems cannot satisfy.
Printed Electronics establish as a key solution to enable such applications.

Printed electronics indicates a set of printing techniques that can realize ultra low-cost~\cite{subramanian2008printed}, large scale~\cite{chen201430} and flexible hardware~\cite{mohammed2017all}.
Printed electronics cannot compete with silicon-based electronics in integration density, area, or speed, mainly due to large feature sizes arising from low-cost and low-resolution printing.
\orange{
The typical operating frequency of printed circuits is from a few Hz to a few kHz~\cite{cadilha2017digital}. 
Similarly, the feature size tends to be several microns~\cite{lei2019low}.
}
Nevertheless, due to its form-factor, conformity, and most importantly ultra-low fabrication costs of the additive manufacturing processes, that enable hardware on-demand even at very low quantities, printed electronics can target application domains untouchable by silicon VLSI.
\orange{However, their large feature sizes and high intrinsic transistor gate capacitances result in elevated power and area compared to nanometer technologies.
}
\orange{Additionally}, the integration density of printed circuits is orders of magnitude lower compared to silicon systems, prohibiting thus the realization of complex circuits.
For example, a printed 
\label{commentR1C4b}MAC \yellow{(multiply–accumulate)} unit --core of most machine learning (ML) circuits-- is six orders of magnitude larger than the CMOS one, \orange{while its power consumption is 8x higher~\cite{Mubarik:MICRO:2020:printedml}.}

% \orange{
% This setting shows the need for more agile tools that avoid general-purpose ML acceleration and instead explore model-specific acceleration methods, especially when targeting embedded systems that are often task-specific.
% }
\orange{
Since low-cost embedded ML systems are usually task-specific and in order to address the aforementioned limitations, there is the opportunity of generating model-specific ML circuits that are more efficient compared to general-purpose ones.}
Leveraging the potential for high customization, offered by the low-fabrication costs of printed circuits,~\cite{Mubarik:MICRO:2020:printedml} 
% designed bespoke ML circuits tailored to specific models.
\label{commentR1C3}\yellow{designed bespoke ML circuits.
The term bespoke refers to fully-customized, even per ML model and dataset, circuit implementations.
% Such customization can be achieved for example by hardwiring the weights in a ML circuit description or by removing instructions from a microprocessor~\cite{Bleier:ISCA:2020:printedmicro}.
}
Still,~\cite{Mubarik:MICRO:2020:printedml} examined only simple ML models (e.g., Decision Tree) rather than complex Multilayer Perceptrons (MLPs) that they pose infeasible hardware overheads for the ultra-resource constrained printed circuits. 
% \orange{
% This hardware overhead puts additional constraints for implementing conventional circuits and thereby limits the works on printed ML designs.
% }
\orange{Thereby, works on printed ML designs are limited.}

Fortunately, approximate computing (AC) can be employed to reduce the associated hardware overheads at the cost of some accuracy loss.
Exploiting the inherent error resilience of a large number of application domains, e.g., ML, AC relaxes the accuracy of some computations and achieves improvements in hardware metrics such as area and power.
\orange{
The authors in~\cite{Armeniakos:DATE2022:axml}, designed for the first time approximate printed ML classifiers.
Through post-training algorithmic weight-approximation and hardware-level gate-pruning, \cite{Armeniakos:DATE2022:axml} generates printed ML circuits with high area and power gains.
However, as in~\cite{Mubarik:MICRO:2020:printedml}, authors in~\cite{Armeniakos:DATE2022:axml} deduced that the hardware overheads of MLPs are prohibitive for printed circuits and additional optimizations are still required.
%In~\cite{Armeniakos:DATE2022:axml}, AC principles and printed electronics are jointly combined to enable different types of ML classifiers in printed technologies.
%Through post-training approximations at algorithmic and hardware level, \cite{Armeniakos:DATE2022:axml} generates printed circuits with significant area and power gains.
%However, it is demonstrated, that enabling more complex ML models such as MLPs with usually higher accuracy, needs further improvement.
}

In this work, we embrace the bespoke design paradigm, enhance it with AC, and propose an automated co-design framework for approximate printed MLP circuits.
% \orange{
% , providing the opportunity for developers to progressively and iteratively design high-quality approximate circuits for resource-constrained printed applications. 
% }
% \orange{
% To further improve the area efficiency compared to state-of-the-art approximate bespoke ML implementations~\cite{DATE}, we exploit the knowledge of datasets for a training-aware approximation technique.
% }
Leveraging the observation that the area of bespoke MLP circuits is intrinsically correlated to the MLP's coefficients, we propose a printing-friendly MLP retraining that selects area-efficient coefficients and achieves $3.3$x lower area, on average, over $10$ MLPs.
By printing-friendly retraining we refer to a bespoke hardware-aware retrain technique, suitable for ultra-resource constrained ML circuits such as our targeted printed ones.
\orange{
Our approach can be seamlessly extended to any printed ML circuit with trainable coefficients.}
\orange{Though}, we focus on MLPs since they constitute the most complex ML classifier for printed applications~\cite{Mubarik:MICRO:2020:printedml}.
Finally,
\orange{in order to further optimize the resource efficiency of our circuits,}
%\orange{in order to replace power-intensive segments of our circuits with inaccurate but simpler components,}
we also approximate the summation of the neuron's products by systematically keeping the most relevant bits and discarding the rest.
\textit{Overall, compared to the exact bespoke baseline, we achieve, on average, $6$x ($5.7$x) area (power) reduction with less than 1\% accuracy loss.}

\noindent
\textbf{Our novel contributions within this work are as follows}:
\begin{enumerate}[topsep=0pt,leftmargin=*]
    \item This is the first work that proposes, implements, and evaluates a software/hardware co-design for printed MLPs while approximating both multiplication and addition operations.
    \item We propose a printing-friendly ML retraining technique.
    \item Our evaluation shows that our framework paves the way, for the first time, towards highly accurate battery powered MLPs, suitable for ultra-resource constrained printed applications\footnote{Our implementations are available at \url{https://github.com/garmeniakos/Ax-Printed-ML-Classifiers}}.
\end{enumerate}

%Printed Electronics are increasingly recognized as the key for the creation of intelligent lightweight electronics, capable of enabling numerous application domains.
%Their flexibility driven by their ultra low cost - in terms of cents - and the low production time (i.e., minutes), make them promising in many common applications such as smart packaging for fast-moving consumer goods (FMCG)~\cite{?}, low-end healthcare (e.g. smart bandages) and radio-frequency Identification (RFID). 
%All of these examples have tight requirements in fabrication costs, which cannot be met by common silicon-based systems and make printed electronics promising candidates.

%A large number of such applications need machine learning (ML) algorithms to perform various classification tasks.
%However, the implementation of ML circuits requires complex designs with numerous operations mainly dominated by multiply-and-accumulates (MACs).
%Unfortunately, in printed technologies the hardware cost of these functions is a major concern as it is extremely larger (i.e. 6 orders of magnitude for a simple MAC~\cite{}) compared to complementary metal-oxide-semiconductors (CMOS).
%This hardware overhead puts additional constraints for implementing conventional circuits and thereby limits the works on printed ML designs.
\section{Related Work}\label{sec:related}

In the past years we have experienced a significant \label{commentR3C3}\yellow{explosion} in the printed electronics research at a variety of application domains.
Targeting applications such as radio-frequency identification (RFID) tags, a pseudo-CMOS logic design for high performance thin-film circuits was presented in~\cite{Myny:2021:dualinput}, 
% while~\cite{Mentens:TCHES2019} was the first work focusing on flexible cryptographic cores and proposed a technique to enhance reliability and prevent malicious circuit modifications.
\green{while in~\cite{Weller:ASPDAC:2020} a 2-input neuron that can be used to support also a MAC operation was fabricated.}

Recently, a flexible 32-bit microprocessor with over $18,000$ gates was fabricated by ARM~\cite{Biggs:Nature2021:flexarm}.
Exploring efficient architectures for printed microprocessors,~\cite{Bleier:ISCA:2020:printedmicro} pruned the ISA 
\label{commentR1C4a}\yellow{(Instruction Set Architecture)} and the respective microarchitecture accordingly and generated tiny printed microprocessors.
However, research on printed ML applications is still at its infancy\yellow{~\cite{Mubarik:MICRO:2020:printedml}, since they are limited only to few neurons implementations rather than more complex ML circuits}.
Acknowledging the need for printed ML inference engines, the authors in \cite{Mubarik:MICRO:2020:printedml} exploited the area-efficiency of bespoke architectures~\cite{Kumar2017Bespoke} and implemented fully customized ML circuits for printed technologies.
Nevertheless, \cite{Mubarik:MICRO:2020:printedml} considered only simple ML models such as Decision Trees and Support Vector Machine Regression due to the high hardware overheads of more complex models such as MLPs.
\green{Moreover,~\cite{Ozer2019Bespoke} described an automated methodology to generate also bespoke classifiers, but no quantification was performed, while further system integration with hardwired machine learning processor for an odour recognition application was fabricated in~\cite{Ozer:Nature:2020}.}
Targeting printed MLPs, \cite{Weller:2021:printed_stoch} employed Stochastic Computing (SC) to reduce their area and power at the cost, however, of a prohibitive accuracy loss in most cases.
\orange{On the other hand,~\cite{Armeniakos:DATE2022:axml} designed approximate MLPs and achieved low accuracy loss but the power reduction was insufficient.}

\orange{
%Apart from the limited work on small ML-driven circuits in printed technology, 
An active research field (also known as TinyML) investigates vigorously resource-efficient ML models which can run on constrained hardware such as IoT-sized devices.
Recently, Google's collaborated CFU Playground~\cite{CFU:Google} proposed a full-stack framework in which bespoke and co-optimized architectures for embedded ML with TinyML focus can be explored.
Although significant speedups can be attained by exploring different custom function units, optimization space for power efficiency has not been comprehensively studied yet.
Furthermore, Kumar et. al.~\cite{Kumar2KBIoT}, targeting low-power devices, developed a tree-based classification algorithm trying to fit in the available memory of such tiny IoT devices.
Similarly, aiming to minimize the required working memory, ProtoNN~\cite{ProtoNN} proposed a k-Nearest Neighbour based algorithm for resource-constrained devices.
However, memory requirements and transfers of both~\cite{Kumar2KBIoT} and~\cite{ProtoNN} consume significant power that acts prohibitively for printed devices.
%Still, new design methodologies are required to overcome restrictions of printed applications and pave the way for adopting energy-harvesting-based solutions~\cite{Mubarik:MICRO:2020:printedml}.
}

Designing approximate arithmetic units such as adders and multipliers has attracted significant research interest~\cite{Honglan:JPROC:2020:axsurv}.
Though, they target non-bespoke arithmetic circuits, unsuitable for ultra resource constrained printed electronics~\cite{Mubarik:MICRO:2020:printedml}.
% \orange{
% Furthermore, despite the large number of approximate adders, research activities on multi-operand adders, as the ones required in MLPs, are negligible.
% }
Finally, vast research focuses on approximate neural network accelerators~ \cite{Mrazek:ICCAD2016:nn,Mrazek2019:ALWANN,Sarwar2018:Alphabet,Zervakis2021:ControlVar,Tasoulas2020:weightoriented}.
However, such deep networks are unrealistic to be considered for printed applications~\cite{Myny:2021:dualinput}.

\orange{
%Table~\ref{tab:related} summarizes and compares the features supported by relevant domains and that are characterized by severe resource-constraints.
%Our work distinguishes from the state-of-the-art and adds a dimension of compatibility for small low-powered devices, by bringing together approximate computing principles and co-design techniques.
%Overall, the proposed workflow offers a step towards realizing complex printed ML circuits with high accuracy that until now seemed unrealizable.
Table~\ref{tab:related} summarizes the above discussion and provides a qualitative comparison of the related works in the field of severe resource-constrainted ML inference.
Although several co-design frameworks have been proposed (e.g., for IoT, FPGAs, TinyML) printed circuits have a much tighter resource constraint and the existing frameworks have not been evaluated on such extreme cases.
Hence, it is highly unclear if existing frameworks will need extremely long time or even fail to find a valid solution.
% Finally, we propose a very fine-grained bespoke design methodology at the level that cannot be achieved in CMOS implementations (e.g. by TinyML).
\label{commentR3C6}\yellow{Finally, we propose a fine-grained design methodology that is bespoke-specific and could also be applied to other technologies that would support such high customizations.} 
Our work distinguishes from most state-of-the-art works and can be only classified along with~\cite{Weller:2021:printed_stoch,Armeniakos:DATE2022:axml} since they all:
i) exploit the efficiency of bespoke implementations,
ii) employ a systematic co-design methodology to boost the efficiency of the generated circuits,
iii) apply non-conventional computing \label{commentR3C1a}(\yellow{Stochastic or Approximate Computing)} to maximize the hardware savings, and 
iv) target the unmatched domain in ultra-limited resources of printed ML applications.
In Section~\ref{sec:results}, we thoroughly evaluate our framework against~\cite{Mubarik:MICRO:2020:printedml,Weller:2021:printed_stoch,Armeniakos:DATE2022:axml}.
In other words, to the best of our knowledge, we compare our work against the only available printed MLP works today.
}

% \begin{table}[t!]
% \renewcommand{\arraystretch}{1.4}
% \begin{tabular}{ccccc}
%  &
%   \begin{tabular}[c]{@{}l@{}}Custom/\\ Bespoke\end{tabular} &
%   \begin{tabular}[c]{@{}l@{}}Approx.\\ Computing\end{tabular} &
%   \begin{tabular}[c]{@{}l@{}}HW/SW\\ Co-Design\end{tabular} &
%   \begin{tabular}[c]{@{}l@{}}End-to-End\\ Solution\end{tabular} \\
% \textbf{Ours}                                                          & \cmark & \cmark & \cmark & \cmark \\
% \cite{Bleier:ISCA:2020:printedmicro,Kumar2017Bespoke} & \cmark & \xmark & \cmark & \cmark \\
% \cite{Mubarik:MICRO:2020:printedml}                   & \cmark & \xmark & \xmark & \cmark \\
% \cite{ProtoNN,Kumar2KBIoT}                            & \xmark & \xmark & \cmark & \xmark \\
% \cite{Weller:2021:printed_stoch}                     & \xmark & $\approx$ & \cmark & \cmark \\
% \cite{CFU:Google}                                     & \cmark & \xmark & \cmark & \xmark
% \end{tabular}
% \end{table}

\begin{table}[t!]
\setlength\tabcolsep{5.3pt}
\renewcommand{\arraystretch}{1.3}
\scriptsize
\caption{Qualitative comparison of related works.}
\begin{threeparttable}
\begin{tabular}{|l|c|c|c|c|c|c|}
\hline
\multicolumn{1}{|c|}{\textbf{Domains}} &
  \textbf{Reference} &
  \textbf{Bespoke\tnote{1}} &
   \textbf{\begin{tabular}[c]{@{}c@{}}AC/\tnote{2}\\ SC\end{tabular}} &
  \textbf{\begin{tabular}[c]{@{}c@{}}HW/SW\tnote{3}\\ Co-Design\end{tabular}} &
%   \textbf{\begin{tabular}[c]{@{}c@{}}End-to-\\ End\end{tabular}} &
  \textbf{\begin{tabular}[c]{@{}c@{}}Battery\tnote{4}\\ Operation\end{tabular}} \\ \hline
  \hline

\multirow{2}{*}{\makecell[l]{Printed \\ Electronics}}  & \cite{Bleier:ISCA:2020:printedmicro,Mubarik:MICRO:2020:printedml,Ozer:Nature:2020} & \cmark & \xmark & \cmark &  \bata \\ \cline{2-6}
& \cite{Weller:2021:printed_stoch,Armeniakos:DATE2022:axml} & \cmark & \cmark & \cmark & \batb \\ \hline
CMOS & \cite{Kumar2017Bespoke} & \cmark & \xmark & \cmark & \xmark \\ \hline
FPGAs & \cite{CFU:Google} & \cmark & \xmark & \cmark & \xmark \\ \hline
TinyML (IoT) & \cite{ProtoNN,Kumar2KBIoT}& \xmark & \xmark & \cmark & \xmark \\ \hline
AC DNN &
%\makecell{\cite{Mrazek:ICCAD2016:nn,Mrazek2019:ALWANN,Sarwar2018:Alphabet} \\ \cite{Zervakis2021:ControlVar,Tasoulas2020:weightoriented}} 
\nocite{Mrazek:ICCAD2016:nn,Mrazek2019:ALWANN,Sarwar2018:Alphabet,Zervakis2021:ControlVar,Tasoulas2020:weightoriented}
\cite{Mrazek:ICCAD2016:nn} --\cite{Tasoulas2020:weightoriented}
& \xmark & \cmark & \cmark & \xmark \\ \hline \hline
PE & \textbf{Ours} & \cmark & \cmark & \cmark &  \batd \\ \hline
\end{tabular}
\begin{tablenotes}\scriptsize
\item[] $^1$Customized implementations.
$^2$AC: Approximate Computing, SC: Stochastic Computing. $^3$HW/SW: Hardware/Software. $^4$Printed batteries $\leq 30$mW
%$^4$ \cite{Bleier:ISCA:2020:printedmicro}~\cite{Mubarik:MICRO:2020:printedml}~$\sim75mW$,\cite{Armeniakos:DATE2022:axml}~\cite{Weller:2021:printed_stoch}~$\sim25mW$, \cite{Kumar2017Bespoke}~\cite{CFU:Google}~$>30mW$, \cite{ProtoNN}~\cite{Kumar2KBIoT}~$\sim200mW$, \cite{Mrazek:ICCAD2016:nn}\cite{Mrazek2019:ALWANN}\cite{Sarwar2018:Alphabet}\cite{Zervakis2021:ControlVar}\cite{Tasoulas2020:weightoriented}~$>1W$, Ours~$\sim8mW$.
\vspace{-2ex}
\end{tablenotes}
\end{threeparttable}
\label{tab:related}
\end{table}
\section{Approximate Printed MLPs}\label{sec:approx}
% \blue{
% \textbf{Length: 2 pages. Figures: 3}
% }
In this section, we present our automated co-design framework for approximate printed MLP circuits.
Briefly, our framework receives as input a trained model (e.g., dumped from scikit-learn) and performs a printing-friendly retraining in which the MLP coefficients are replaced with more area-efficient ones.
Next, our framework generates the Verilog RTL description of the respective bespoke MLP circuit and approximates the summation operation of each neuron by systematically dropping the less significant bits of the summands (i.e., products inputs by coefficients).
Finally, our framework, runs a full search design space exploration (DSE), in which all the approximate circuits are synthesized using EDA tools and the printed \label{commentR3C1c}\yellow{Process Design Kit (PDK)}, to obtain the Pareto-optimal approximate MLPs.

\subsection{Bespoke MLPs}
\label{sec:bespoke}

\begin{table}[t!]
\setlength\tabcolsep{2pt}
\renewcommand{\arraystretch}{1.5}
\caption{Evaluation of Bespoke Printed MLPs. Most MLPs feature unsustainable hardware overheads.}
\label{tab:baselines}
\footnotesize
\centering

\hspace{-9mm}
\begin{tabular}{lcccccc@{\hspace{-4ex}}l@{\hspace{-6ex}}}
\hline
%\textbf{Dataset} & \textbf{Topology}  & \textbf{\#MACs} & \textbf{Accuracy}  & \begin{tabular}[c]{@{}c@{}}\textbf{Area} \textbf{[cm$^\mathbf{2}$]}\end{tabular} & \begin{tabular}[c]{@{}c@{}}\textbf{Power} \textbf{[mW]}\end{tabular} \\ \hline
\textbf{Dataset} & \textbf{Topology}  & \textbf{\#MACs} & \begin{tabular}[c]{@{}c@{}}\textbf{\yellow{Cpd}} \\ \textbf{\yellow{[ms]}}\end{tabular} & \textbf{Acc}  & \begin{tabular}[c]{@{}c@{}}\textbf{Area} \\ \textbf{[cm$^2$]}\end{tabular} & \begin{tabular}[c]{@{}c@{}}\textbf{Power}\\ \textbf{[mW]}\end{tabular} \\ \hline
WhiteWine (WW) & (11,4,7) & 72 & 198 & 0.54 & 31 & 98   & \multirow{8}{*}{\color{BrickRed}$\left.\begin{array}{l}
    \\
    \\
    \\
    \\
    \\
    \\
    \\
    \\
    \end{array}\right\rbrace\rotatebox[origin=c]{90}{\parbox{20mm}{\centering No Adequate\\Power Supply}}$} \\
Cardio (CA) & (21,3,3) &  72 & 199 & 0.88 & 33 & 97      \\
RedWine (RW) & (11,2,6) & 34  & 199 & 0.56 & 18 & 53      \\
Pendigits (PD) & (16,5,10) &  130 & 201 & 0.94 & 67 & 213    \\
Vertebral Col. 3C (V3) & (6,3,3) & 27 & 200  & 0.83 & 8.9 & 36      \\
Balance Scale (BS) & (4,3,3) &21 & 199 & 0.91 & 9.3 & 36      \\
Seeds (SE) & (7,3,3) &30 & 200 & 0.94 & 9.9 & 41 \\
Breast Cancer (BC) & (9,3,2) &33 & 188 & 0.98 & 12      & 40      \\
Vertebral Col. 2C (V2) & (6,3,2) &24 & 114 & 0.90 & 3.5       & 13 & \multirow{2}{*}{\color{RoyalBlue}$\left.\begin{array}{l}
\\
\\
\end{array}\right\rbrace\rotatebox[origin=c]{90}{\parbox{7mm}{\centering Printed\\ Battery\\ Operation}}$} \\
Mammographic (MA) & (5,3,2) & 21  & 197 & 0.86 & 6.8       & 27      \\
\hline
\end{tabular}

\end{table}

The ultra low-cost and on-demand in-situ fabrication in printed electronics enables bespoke implementations, tailored to specific dataset or usecase, that enable circuits with phenomenal area reduction compared to the respective conventional general purpose ones~\cite{Mubarik:MICRO:2020:printedml}.
% As a result, bespoke implementations prevail as the most prominent solution to realize ultra-resource constrained printed circuits.
\orange{Leveraging the ultra low-cost and on-demand in-situ fabrication, bespoke implementations prevail as the most prominent solution to realize ultra-resource constrained printed circuits~\cite{Mubarik:MICRO:2020:printedml,Bleier:ISCA:2020:printedmicro}.}
Driven by this potential, we also embrace the bespoke design paradigm for printed MLPs.
In such highly customized circuits, the coefficients of the ML model (MLPs in our case) are hardwired in the circuit implementation itself~\cite{Mubarik:MICRO:2020:printedml}.

For our analysis, we consider $10$ MLP classifiers (see Table~\ref{tab:baselines}) trained on varying datasets of the UCI ML repository~\cite{Dua:2019:uci}.
These datasets are selected similarly to~\cite{Mubarik:MICRO:2020:printedml} and \cite{Weller:2021:printed_stoch}.
To train the MLPs, scikit-learn and the randomized parameter optimization with $5$-fold cross validation are used.
Inputs are normalized to $[0,1]$ and we use a random $70$\%/$30$\% train/test set split.
The topology of the MLPs is $\#input\!\times\!L\!\times\!\#output$ with $L\leq5$ and the ReLU is used for activation function.
$L$ is selected so that each MLP achieves close to maximum accuracy while $L$ (i.e., the number of its hidden nodes) is minimized.
At the final stage an argmax function is used to translate and map numerical predictions (i.e., values of output neurons) to a class, from which classification accuracy can be obtained.

Table~\ref{tab:baselines} presents the evaluation of the examined MLP circuits.
All circuits in Table~\ref{tab:baselines} are implemented following the bespoke fully-parallel (i.e., $1$ inference/cycle) state-of-the-art design methodology of~\cite{Mubarik:MICRO:2020:printedml}.
%\orange{Since DFFs in printed technologies are considerable more expensive than combinational cells, the best printed architecture, which we also follow in our designs, is the single-stage pipeline~\cite{Bleier:ISCA:2020:printedmicro}.}
In the remainder, these circuits will be referred to as baseline circuits.
Fixed-point arithmetic is used with $4$ bits for the inputs and $8$ bits for coefficients, achieving close to floating-point accuracy.
Nevertheless, since coefficients are hardwired in the circuit, we use the bare-minimum precision for each coefficient (e.g., ``$3$'' uses only $2$ bits).
Circuit synthesis is performed with Synopsys Design Compiler and targeting the Electrolyte Gated Transistor (EGT) library~\cite{Bleier:ISCA:2020:printedmicro}, which is a low-voltage inkjet-printed technology that allows battery powered printed circuits.
\label{commentR2C2a}\yellow{
Since our objective in our optimization problem is to further improve the area efficiency rather than the performance, all MLPs are synthesized and simulated at relaxed timing constraints, i.e., $250$ms per inference for Pendigits and $200$ms for the rest ones.
}
% To further improve the area efficiency, all MLPs are synthesized at relaxed timing constraints, i.e., $250$ms per inference for Pendigits and $200$ms for the rest ones.
These performance values comply with typical operating frequencies in printed electronics~\cite{cadilha2017digital}, \yellow{while delay gains due to our approximations are presented in Section~\ref{sec:result1}.}
Power estimation is performed using Synopsys PrimeTime and switching activity obtained from circuit simulations with Questasim.
% The aforementioned tool-flow is used in all our hardware evaluations.

\yellow{Due to the nature of printed applications, they all pose tight area requirements.
As a rule of thumb and similar to~\cite{Kumar2017Bespoke}, we consider the $10cm^2$ and $30mW$ (i.e., maximum power of a single printed battery) as a hard constraint for most printed applications.
As shown in Table~\ref{tab:baselines} the average area of the MLP circuits is prohibitive for such applications~\cite{Kumar2017Bespoke}.}~\label{commentR1C1}
Moreover, only the Vertebral Column 2C and the Mammographic MLPs can be powered by an existing printed battery (e.g., Molex $30$mW battery) while for the rest MLPs it doesn't exist any adequate power supply~\cite{Mubarik:MICRO:2020:printedml}.

\subsection{Printing-Friendly MLP Retraining}\label{subsec:retrain}
% \blue{
% \textbf{Length: 0.75 page. Figures: 1}
% \begin{itemize}
%     \item Goal: obtain printed-friendly coefficients $\rightarrow$ reduce the size of the multipliers.
%     \item Describe weight clustering based on the area of the multiplier
%     \item 1's complement for positive/negative weights
%     \item Describe our retrain methodology (5 steps+score function)
%     \item overall methodology figure
% \end{itemize}
% }
% \begin{figure}
%     \centering{\phantomsubcaption\label{fig:neuronmc}\phantomsubcaption\label{fig:mult_area}\phantomsubcaption\label{fig:cluster_box}}
%     \includegraphics[width=1\columnwidth]{graphs/area+boxplot+monte_carlo(1).PDF}
%     \caption{a) $1000$-point Monte Carlo analysis of the area of bespoke neurons w.r.t. the values of the coefficients. b) Area of bespoke printed multipliers with $4$-bit inputs and coefficients in [$-128$,$127$]. For reference the area of the conventional $4\times 8$ multiplier is 83.61mm$^{2}$. c) Area analysis of the clustered coefficients for $4$-bit inputs and coefficients in [0,127]. }
%     \label{fig:mult_area_box}
% \end{figure}

\begin{figure}[t!]
\centering
\includegraphics{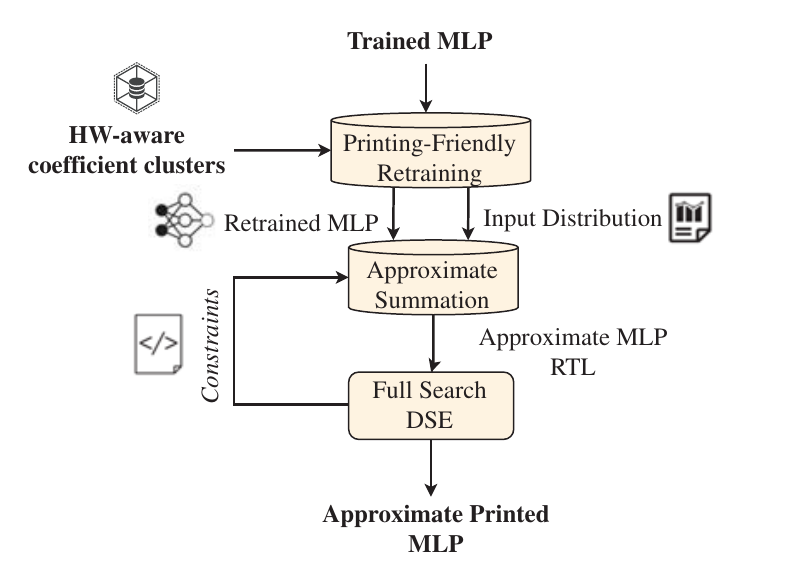}
\caption{
Abstract overview of our framework.}
\label{fig:frame}
\end{figure}

\begin{figure}[t!]
\centering{\phantomsubcaption\label{fig:neuronmc}\phantomsubcaption\label{fig:mult_area}}
\centering
\includegraphics{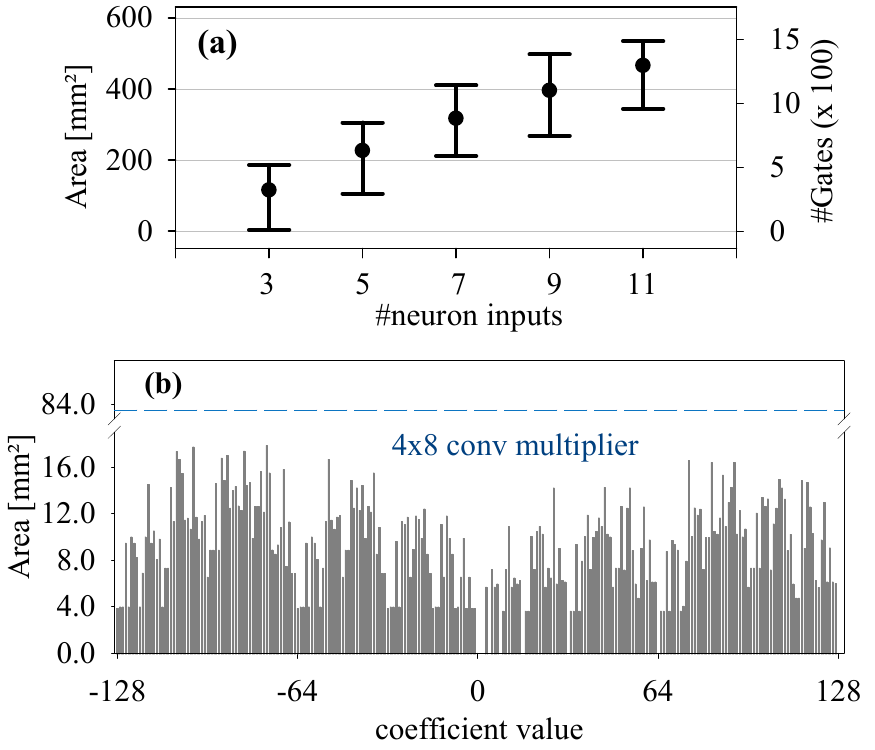}
\caption{
a) $1000$-point Monte Carlo analysis of the area of bespoke neurons w.r.t. the values of the coefficients.
b) Area of bespoke printed multipliers with $4$-bit inputs and coefficients in [$-128$,$127$].}
\label{fig:test}
\end{figure}

Our framework (Fig.~\ref{fig:frame}) enhances bespoke circuits with approximate computing to further improve the hardware efficiency of the former.
Although bespoke circuits form the most promising solution for printed ML applications, bespoke implementations open new rooms for optimizations in printed circuits that are barely explored up to now.
For example, in Fig.~\ref{fig:neuronmc} we perform a $1000$-point Monte Carlo analysis of the area of bespoke neurons w.r.t. the values of the coefficients.
Neurons are the building block of MLPs.
As shown, irrespectively of the size of the neuron (\#inputs), the area of the neuron features very high variation.
For example, in Fig.~\ref{fig:neuronmc} the average standard deviation is 63mm$^2$ or else 175 gates.
Therefore, there is a high potential, with a proper coefficient selection, to keep the bespoke neuron's area minimal.
We further investigate this and present in Fig.~\ref{fig:mult_area} the area of bespoke multipliers $a\cdot w$.
Multipliers constitute the core components of a neuron.
The value $w$ is hardwired in each bespoke multiplier, the input $a$ is $4$ bits and the coefficient $w$ is up to $8$ bits, i.e., $w \in [-128,127]$.
It is noteworthy that the area of the bespoke multipliers is more than $5$x lower than the area of the respective $4\times 8$ conventional multiplier.
From Fig.~\ref{fig:mult_area}, it is evident that there is an intrinsic correlation between the value of the coefficient $w$ and the area of the bespoke multiplier and consequently, the area of the neuron.
Importantly, in the cases that the coefficient is a power of two, the multiplier's area is nullified \orange{(i.e., multiplier is replaced by only simple wiring)}.

%Motivated by the above observations we leverage our hardware knowledge from our bespoke architectures to implement a printed-friendly retraining.
Motivated by the above observations we leverage our bespoke hardware analysis to implement a printing-friendly retraining.
Our main goal is to replace the coefficients of the MLP with more area-efficient ones and as a result, minimize the area of the required multipliers while maintaining high accuracy in the meantime.

\begin{figure}[t!]
\centering
\includegraphics{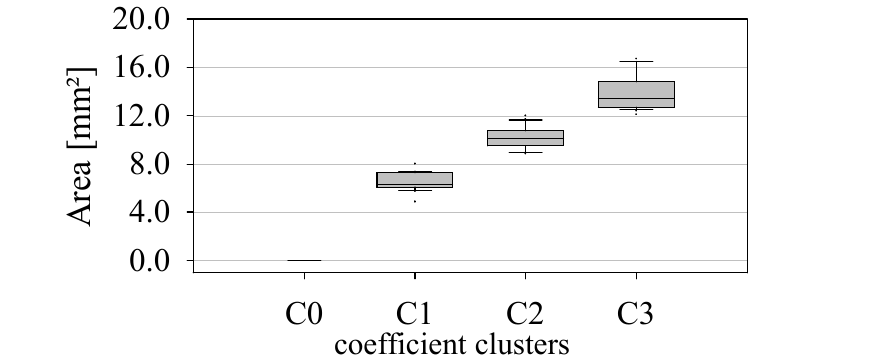}
\caption{
Area analysis of the clustered coefficients for $4$-bit inputs and coefficients in [0,127].}
\label{fig:cluster_box}
\end{figure}

As a step towards enabling printing-friendly retraining, we need to distinguish the coefficients based on their area-efficiency.
To achieve this, we use K-means and cluster the coefficients with respect to the area of the respective bespoke multiplier.
Without loss of generality, we consider up to $8$ bits for the coefficients and we cluster the coefficients into four groups $C_0$-$C_3$.
Group $C_0$ comprises only powers of two, resulting thus to zero-area multipliers, while the multipliers generated by the coefficients of $C_i$ feature larger area than the multipliers generated by the coefficients of $C_j$ 
\label{commentR2C4}if \yellow{$i>j$}.
For example, we clustered the coefficients $w$, $\forall w \in [0, 127]$, considering bespoke multipliers with $4$ bits input.
Fig.~\ref{fig:cluster_box} presents the area of the bespoke multipliers of each cluster.
As shown in Fig.~\ref{fig:mult_area}, the negative coefficients produce multipliers with larger area than the respective positive ones.
Nevertheless, as we will explain later, during retraining we assume that the positive and negative coefficient multipliers feature the same area.
For this reason, we perform the clustering only for the positive coefficients.
Moreover, we clustered the coefficients using several input sizes from $4$ up to $16$ bits and obtained identical results.
The latter is explained by the fact that increasing the input size impacts all the bespoke multipliers similarly, irrespectively of the coefficient value.
Therefore, although the neurons of the hidden and output layers might feature different input sizes, the same clustering can be used.
Finally, to cluster the coefficients, we need to synthesize (once for all MLPs) all the positive bespoke multipliers.
In our case, for $128$ bespoke multipliers, it required less than a minute using $10$ threads \label{commentR3C8}\yellow{in a Xeon E5-2650 server with 64-GB RAM.}

Algorithm~\ref{alg:cap} presents an abstract overview of our printing-friendly MLP retraining.
Our framework receives as inputs a trained model, the training dataset, and a user-defined accuracy loss threshold \label{commentR3C9}\yellow{that remains constant throughout the retraining process and could be relaxed for higher potential gains.
% As a next step and for higher potential gains this threshold could be relaxed.
}
\label{commentR2C1}\yellow{Note that accuracy loss is defined as: $Accuracy_{exact}-Accuracy_{approx.}$.}
Given an already trained MLP with fixed topology (namely MLP0), our algorithm initiates our MLP$^\prime$ to MLP0 and retrains MLP$^\prime$ trying to assign all its coefficients to $C_0$ ($VC \leftarrow \{w,-w:w \in C_0\}$).
If after $m$ epochs the accuracy is below a given threshold then i) we reset the training (MLP$^\prime$ $\leftarrow$ MLP0), ii) we increase the number of available values for the coefficients by gradually considering more clusters, and iii) we repeat the training for another $m$ epochs each time.
A solution always exists since at the worst case all the coefficient clusters will be consumed.
During the $m$ epochs retraining, if the accuracy is unacceptable and no coefficients are updated, we increase the learning rate to allow jumps~\cite{Sarwar2018:Alphabet}.
The latter is crucial since if the distance of the available values in $VC$ is large, then with a small learning rate the coefficients might always be mapped to the same value in $VC$.
Finally, to enable area-awareness during training we add a small penalty at the obtained accuracy based on the area of the required bespoke multipliers.
\orange{
This is crucial, when using more than one clusters, in order to guide retraining towards selecting more coefficients from the lower (more area efficient) clusters.
}
To achieve this, we calculate the following score function:

\begin{equation}\label{eq:score}
\begin{split}
     S =\, & \alpha \cdot \frac{\text{accuracy(MLP$^\prime$)}}{\text{accuracy(MLP0)}} \\
     &+ (1-a) \times \frac{\mathrm{AR}(\text{MLP0})-\mathrm{AR}(\text{MLP$^\prime$})}{\mathrm{AR}(\text{MLP0})},
\end{split}
\end{equation}
where $\mathrm{AR}$ is the sum of the area of the bespoke multipliers instantiated by each MLP.
These area values are calculated based on the input sizes of each neuron and are stored in a look-up table to be used during retraining.
Again the time required is negligible ($5$min at the worst case examined) and for the negative coefficients we use the area of the respective positive ones.
When MLP$^\prime$ and MLP0 feature the same accuracy and MLP$^\prime$ uses only $C_0$ (only power of two), $\mathrm{AR}(\text{MLP$^\prime$})$ becomes zero and \eqref{eq:score} takes its maximum value, i.e.,~$S=1$.
When MLP$^\prime$ and MLP0 are the same (e.g., initial assignment), $S$ in \eqref{eq:score} equals $\alpha$.
In our work, targeting high accuracy printed MLPs, we set $\alpha=0.8$.
Nevertheless, the area-accuracy tradeoff w.r.t. $\alpha$ needs to be explored more comprehensively in the future.
Finally, we set the retraining epochs $m=10$ to constraint the execution time spend in retraining (i.e., $40$ epochs at most).
On average, 4min were required by our printing-friendly retraining.

\begin{algorithm}[t!]
\caption{Printing-Friendly Retraining Pseudocode}\label{alg:cap}
\small
\textbf{Input:} 1) Trained Model: MLP0, 2) Accuracy Loss Threshold: $T$, \\ \hspace{8mm} 3) Train Dataset\\
\textbf{Output:} 1) Printing-Friendly Model: MLP$^\prime$\\
\begin{algorithmic}[1]
\State $C_i, \forall i \leftarrow$ Cluster Coefficients
\State $VC  \leftarrow \{\}$
\State \textbf{for} $0 \leq i <$\#Clusters
\State \hspace{2mm} $VC  \leftarrow VC \cup \{w,-w:w \in C_i\}$
\State \hspace{2mm} MLP$^\prime$ $\leftarrow$ MLP0
\State \hspace{2mm} retrain MLP$^\prime$ for $m$ epochs
\begin{itemize}\setlength{\itemindent}{+3mm}
    \item feedforward: evaluate score function Eq. \eqref{eq:score}
    \item coefficient update: convert coefficients to fixed point \& \newline
    map each coefficient to its closest value in $VC$
    \item adjust learning: if no coefficient updated $\rightarrow$ increase learning rate
\end{itemize}
\State \hspace{2mm} \textbf{if} (accuracy(MLP$^\prime$) $\geq$ accuracy(MLP0) $- T$) \textbf{break}
\State \textbf{return} MLP$^\prime$
\end{algorithmic}
\end{algorithm}

Overall, our coefficient cluster-based retraining approach constraints the search space of printing-friendly coefficients and helps exploring early and with high confidence area-efficient solutions.
As more clusters are gradually used in retraining, the penalty imposed by our score function guides the training algorithm to limit the number of coefficients selected from the higher clusters and instead select more area-efficient coefficients from the lower clusters.

\subsection{Approximate Bespoke Neuron}\label{subsec:axsum}
\begin{figure}[t!]
    \centering
    \includegraphics[width=1\columnwidth]{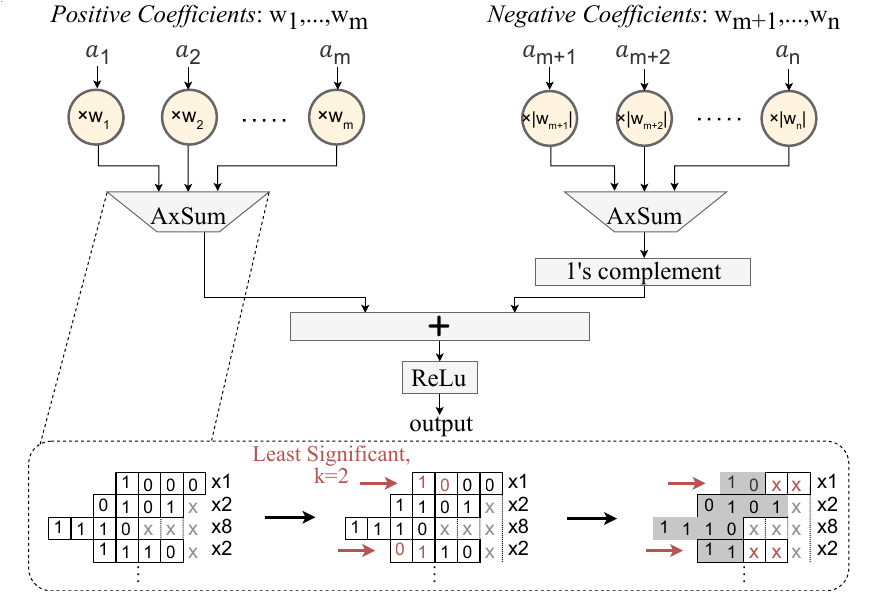}
    \caption{Overview of our approximate bespoke neuron.}
    \label{fig:axsum}
\end{figure}

\blue{
\textbf{Length: 1 page. Figures: 1}
\begin{itemize}
    \item motivate shifted summands, e.g., different powers of two, (figure)
    \item keep only the MSBs for the less significant summands 
    \item significance function
    \item methodology + respective DSE
\end{itemize}
}
Each neuron calculates a weighted sum: \\
\begin{equation}\label{eq:wsum}
    S=\sum_{\forall i}{a_i\cdot w_i},
\end{equation}
where $w_i$ are the neuron's coefficients (or weights) and $a_i$ are its inputs.
After retraining, the coefficients of the MLP are replaced with more printing-friendly ones, leading to reduced hardware requirements w.r.t. the neuron's bespoke multipliers.
Still, the summation of the products of the bespoke multipliers results in considerable hardware overhead.
Fig.~\ref{fig:axsum} depicts the implementation of our approximate neuron.
Overall, we calculate the following: \\
\begin{equation}\label{eq:axwsum}
    S^\prime=\underbrace{\sum_{\forall i: w_i>=0}{a_i\cdot w_i}}_{S_p}+\Big(\sim\!\underbrace{\sum_{\forall i: w_i<0}{a_i\cdot |w_i|}}_{S_n}\Big),
\end{equation}
where $\sim$ refers to logical NOT (i.e., 1's complement).
Exploiting that the inputs of each neuron are positive (see Section~\ref{sec:bespoke}), we know apriori the sign of each product (i.e., same with the sign of the corresponding coefficient). 
Hence, we split the coefficients into positives and negatives and for the negative ones we use their absolute value to generate their bespoke multipliers.
A different adder tree is used to sum each set of products and 1's complement (instead of 2's complement) is used to negate the sum of the negative coefficients ($S_n$).
\orange{Finally, $S_p$ and $\sim S_n$ are added together.}
Calculating~\eqref{eq:axwsum} requires only positive bespoke multipliers that feature significantly lower area than the negative ones.
Moreover, with~\eqref{eq:axwsum} we eliminate many full adders that would be required just for sign extension.
If the neuron doesn't have any negative coefficients, the right parts of~\eqref{eq:axwsum} and of Fig.~\ref{fig:axsum} are omitted.
\orange{Again, such high customization is feasible only in bespoke circuits.
}

To further reduce the area complexity, we approximate the adders that produce the sums $S_p$ and $S_n$.
To achieve this, we keep only some MSBs of the least significant summands (products $a_i \cdot w_i$) and we discard the rest.
Hence, the more summands are approximated, the higher area reduction is achieved.
An illustrative example is depicted in Fig.~\ref{fig:axsum}.
Motivated by our cluster-based retraining, many coefficients are assigned to a power of two.
Intuitively, inputs multiplied by high powers of two will generate considerably more significant products for the final result compared to inputs multiplied by small powers of two.
However, despite this intuitive observation the significance of the product $a_i\cdot w_i$ depends also on of $a_i$.
Thus, we define the significance of each product as follows:

\begin{equation}\label{eq:sig}
    G_i=|w_i\frac{\mathrm{E}[a_i]}{\sum_{\forall i}{\big(\mathrm{E}[a_i]\cdot w_i \big)}}|.
\end{equation}
In other words,~\eqref{eq:sig} calculates the ratio of the average value of each product $a_i\cdot w_i$ over the average sum of the neuron.
For each neuron, $G_i, \forall i$, is easily calculated by just capturing the inputs distribution during training.
Exploiting this high-level information, we approximate accordingly, at design time, the summation operations ($S_p$ and $S_n$).
For each product $a_i\cdot w_i$, if $G_i$ is less or equal to a given threshold $G$, we keep only the $k$ MSBs with $k\in [1,3]$.
Hence, for each neuron, the approximate sum (AxSum) is given by:
\begin{equation}
\begin{gathered}\label{eq:axwsumk}
S^\prime=S_p+(\sim S_n)\\ \\
S_p=\quad\sum_{\mathclap{\substack{\forall i: w_i>=0,\\G_i>G}}}{p_i}\,+\quad\sum_{\mathclap{\substack{\forall i: w_i>=0,\\G_i\leq G}}}{p_i[n_i-1:n_i-k]2^{n_i-k}},\\ \\
S_n=\quad\sum_{\mathclap{\substack{\forall i: w_i<0,\\G_i>G}}}{p_i}\,+\quad\sum_{\mathclap{\substack{\forall i: w_i<0,\\G_i\leq G}}}{p_i[n_i-1:n_i-k]2^{n_i-k}},\\ \\
p_i=a_i \cdot |w_i|,\quad n_i=\$size(|w_i|)+\$size(a_i), \forall i
\end{gathered}
\end{equation}
where $n_i, \forall i$, is the size of each product, e.g., for $w_i$=\textpm$7$ and $4$-bit inputs, $n_i$ is $7$ bits.
Note that, \eqref{eq:axwsumk} refers to each neuron, i.e., different neurons might feature different $k$ and $G$ values.
To reduce the size of the design space, we consider the same $k$ value for all neurons and one $G$ value per layer.
The latter is based on the fact that different layers feature different sensitivity to approximation~\cite{Tasoulas2020:weightoriented}.

\begin{figure}
    \centering
    \includegraphics{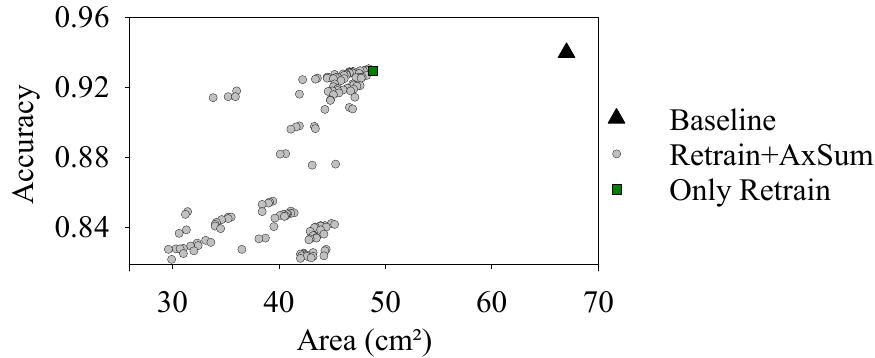}
    \caption{Accuracy-Area Pareto space of the PD MLP.}
    \label{fig:dse}
\end{figure}

\begin{figure*}
    \centering
    \includegraphics{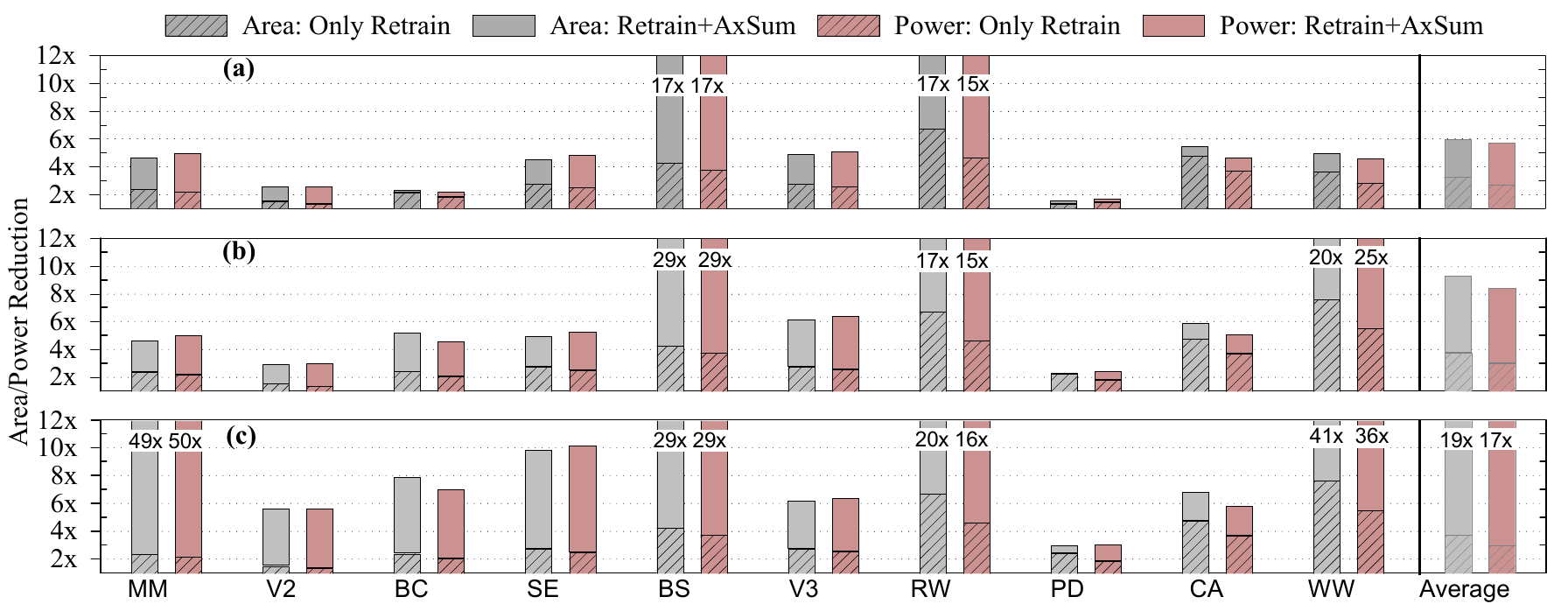}
    \caption{Area and power reduction our proposed approximate MLPs compared to the exact state-of-the-art bespoke baseline~\cite{Mubarik:MICRO:2020:printedml}. Three accuracy loss thresholds $T$ are examined for our approach: a) 1\%, b) 2\%, and c) 5\%.}\vspace{-1ex}
    \label{fig:barplots}
\end{figure*}

Finally, we perform an exhaustive DSE w.r.t. the value $k \in [1,3]$ and all the possible values of $G$ for each layer.
Each design point is synthesized and simulated and a Pareto analysis is performed to obtain the Area-Accuracy Pareto Front.
Unlike conventional silicon VLSI, in printed electronics the examined ML models are rather small in size.
Hence, synthesis and simulation of each design requires a couple of minutes at most.
Thus, we can obtain fast enough the Pareto optimal designs through an exhaustive DSE.
On average, DSE required only $7$ min using $10$ threads (i.e., limit of our available EDA licenses).
\orange{At the worst case, the DSE required $1$h for the PD MLP, which is however very large to be considered for a printed application.}
For example, Fig.~\ref{fig:dse} presents the Accuracy-Area Pareto space for the PD MLP and up to $20$\% accuracy loss.
\yellow{\label{commentR1C2}The green square is the design that applies only our printing-friendly retraining, while gray circles are approximate derivatives of the green square and are derived from different approximate configurations of $k$ and $G$.
% (i.e., approximate configurations).
Overall, the designs generated using ``Retrain$+$AxSum" constitute the Pareto front and achieve 2x area reduction 
for only $2\%$ accuracy loss.
}
% As shown, $2$x area reduction is achieved for only 2\% accuracy loss.

Note that in~\eqref{eq:wsum}-\eqref{eq:axwsumk} we didn't consider a neuron bias term to simplify the presented analysis.
If a neuron has a bias, then similarly to the coefficients, the bias is hardwired in the circuit itself.
Moreover, if the bias is positive, then it is added in $S_p$ among with the positive products.
Similarly, if it is negative its absolute value is added in $S_n$.

\section{Results}\label{sec:results}

In this section, we investigate the effectiveness of our proposed co-design framework in enabling printed MLPs with minimal accuracy loss.
We evaluate the area, power, and accuracy of our approximate MLPs against the state-of-the-art exact bespoke circuits~\cite{Mubarik:MICRO:2020:printedml} (see Table~\ref{tab:baselines}) and we also compare our framework against the stochastic MLPs~\cite{Weller:2021:printed_stoch, Armeniakos:DATE2022:axml} that also trade accuracy for area and power gains.
\orange{To the best of our knowledge,~\cite{Mubarik:MICRO:2020:printedml,Weller:2021:printed_stoch, Armeniakos:DATE2022:axml} are the only available works on printed ML circuits.
Overall, in our AxSum DSEs, we evaluated more than 600 circuits in order to extract the optimal approximate printed MLPs.
Note that the accuracy reported in this section is on the test dataset while all our optimizations are performed on the train dataset.}

\subsection{Comparison With Exact Baseline}\label{sec:result1}
\begin{figure}[t!]
    \centering
    \includegraphics{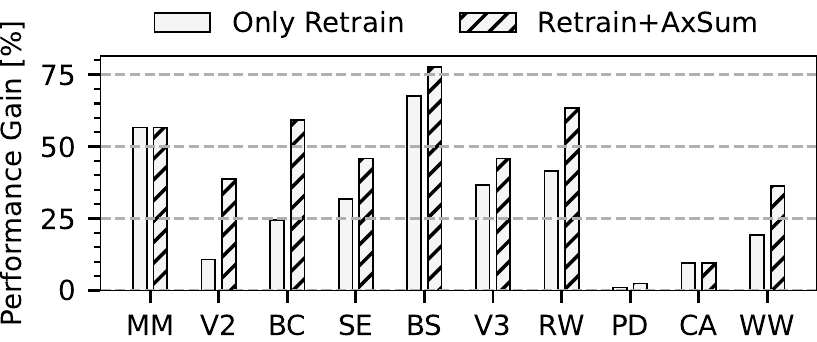}
    \caption{\yellow{Critical path delay gains for our approximate MLPs compared to the exact bespoke baseline~\cite{Mubarik:MICRO:2020:printedml} and for 1\% accuracy loss threshold.}}
    \label{fig:cpd}
\end{figure}

Fig.~\ref{fig:barplots} presents the area and power gains of our framework (``Retrain+AxSum'') w.r.t. the exact bespoke circuit~\cite{Mubarik:MICRO:2020:printedml}.
Both our MLPs and~\cite{Mubarik:MICRO:2020:printedml} are synthesized with the same timing constraints.
In Fig.~\ref{fig:barplots}, we consider three accuracy loss thresholds: $1\%$, $2\%$ and $5\%$.
For each threshold, we selected the Pareto-optimal MLP from our DSE that satisfies it.
In addition, we also report the gains obtained when applying only our printing-friendly retraining (``Only Retrain'').
The examined thresholds refer to the overall accuracy loss, i.e., due to printing-friendly retrain and AxSum.
However, since multipliers consume the most area and power in MLPs~\cite{Mrazek:ICCAD2016:nn}, we assign all the available accuracy loss budget to our retraining algorithm.
Then, if there is still room for further approximation, we apply our AxSum.
Though, as Fig.~\ref{fig:barplots} shows, AxSum is always used.
Compared to~\cite{Mubarik:MICRO:2020:printedml}, our framework delivers very significant area and power gains.
Specifically, we achieve $6.0$x ($5.7$x), $9.3$x ($8.4$x), and $19.2$x ($17.4$x) lower area (power), for up to $1\%$, $2\%$ and $5\%$ accuracy loss, respectively.
The corresponding values when using only our printing friendly retraining are $3.30$x ($2.72$x), $3.78$x ($3.03$x), and $3.80$x ($3.04$x).
Therefore, both our retrain as well as our AxSum methods contribute significantly towards the final area and power gains.
Nevertheless, the area and power savings of retraining saturate after $2$\% accuracy loss.
Hence, above $2$\%, further gains are subject only to our AxSum.
% This highlights the importance of our approximate summation on top of our printing-friendly retraining.
For $1$\% accuracy loss, after retraining, \yellow{WhiteWine and BreastCancer} MLPs used the first two coefficient clusters, \yellow{Pendigits} used all the clusters, while the rest MLPs used only $C_0$ (i.e., maximum area reduction w.r.t. multipliers). \label{commentR3C1b}
Apart from PD, all MLPs used only $C_0$ for $2$\% and $5$\% accuracy loss.
For $2$\%, PD used the first three clusters, while for $5$\% PD used the first two.
Since for $2$\% and $5$\% almost all MLPs were retrained only with coefficients from $C_0$, our printing-friendly retraining achieved the maximum possible savings and thus, this explains why its average area/power gains saturate after $2$\%.
Overall, the PD MLP elucidates the different aspects of our retraining methodology.
For tight accuracy constraints, all the coefficient clusters are used and our score function still ensures hardware gains, while for relaxed constraints less clusters are needed.
It is noteworthy, that the baseline PD MLP uses only $21$ coefficients from $C_0$, while for $1$\% accuracy loss, the score function~\eqref{eq:score} forced our retraining to select $64$ coefficients from $C_0$.
This explains why our ``Only Retrain'' PD MLP achieves $1.75$x lower area than~\cite{Mubarik:MICRO:2020:printedml}, although both of them use coefficients from all the clusters (i.e., impact of our score function).
\yellow{
Finally, applying both our retraining and AxSum approximation opens the road also for higher performance.
Fig.~\ref{fig:cpd} illustrates the delay reduction for each model, when applying our proposed approximations, compared to their exact (baseline) designs.
On average, 44\% CPD reduction is achieved for less than 1\% accuracy loss.
}

\begin{figure}[t!]
    \centering
    \includegraphics{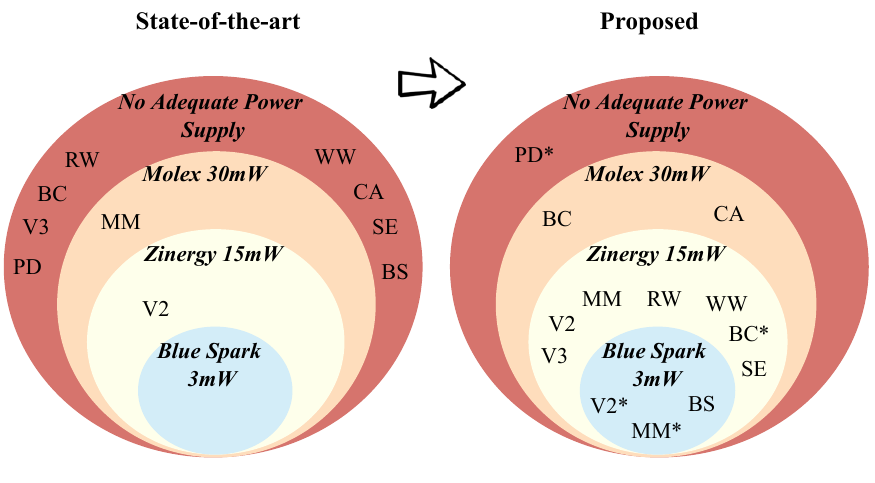}
    \caption{Power supply classification of printed MLPs w.r.t. existing printed batteries. a) State-of-the-art bespoke MLPs~\cite{Mubarik:MICRO:2020:printedml}, b) ours. All our MLPs feature up to $1$\% accuracy loss, except the $^*$denoted ones that feature $5$\%.}
    \label{fig:batteries}
\end{figure}

Fig.~\ref{fig:batteries} quantifies in an illustrative manner the tangible impact of our work.
In Fig.~\ref{fig:batteries}, up to $5$\% accuracy loss is considered.
As shown, the current state-of-the-art cannot support printed MLPs.
For example, for $8/10$ MLPs there is no existing adequate power supply.
On the other hand, our framework enables $9/10$ battery powered printed MLPs.
Most of the MLPs can now be powered by only a Zinergy $15$mW battery, while $3/10$ can use a Blue Spark $3$mW battery.
\orange{Similarly, significant gains in area can now enable the practical and realistic printed applications(see unattainable area of the exact designs in Table~\ref{tab:baselines}).}
\textit{This huge shift on supported printed MLPs may open new horizons for the realization of smart printed applications.}

% \begin{table}[t!]
% \setlength\tabcolsep{4pt}
% \renewcommand{\arraystretch}{1.5}
% \caption{cpd}
% \label{tab:cpd}
% \begin{tabular}{ccccccccccc}
% \hline \hline
%                                                                                        & \multicolumn{10}{c}{\textbf{Critical Path Delay (ms)}}    \\ \hline
% \multicolumn{1}{c|}{} & \textbf{MM} & \textbf{V2} & \textbf{BC} & \textbf{SE} & \textbf{BS} & \textbf{V3} & \textbf{RW} & \textbf{PD} & \textbf{CA} & \textbf{WW} \\  \hline
% \multicolumn{1}{c|}{\textbf{Baseline}}                                                 & 197 & 114 & 188 & 200 & 198 & 200 & 199 & 201 & 199 & 199 \\ \hline
% \multicolumn{1}{c|}{\textbf{\begin{tabular}[c]{@{}c@{}}Only\\ Retrain\end{tabular}}}   & 85  & 102 & 142 & 136 & 64  & 126 & 117 & 199 & 180 & 161 \\ \hline
% \multicolumn{1}{c|}{\textbf{\begin{tabular}[c]{@{}c@{}}Retrain\\ +AxSum\end{tabular}}} & 85  & 70  & 77  & 108 & 44  & 108 & 73  & 196 & 180 & 127 \\ \hline \hline
% \end{tabular}
% \end{table}

\subsection{Comparison With AC- and SC- based Printed MLPs}

\begin{figure}[t!]
    \centering
    \includegraphics{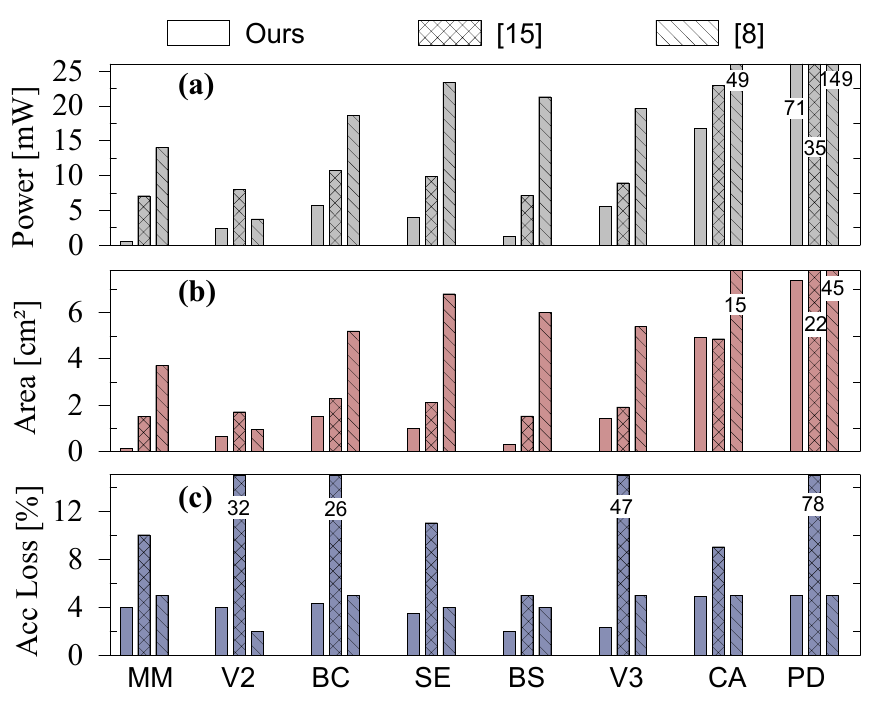}
    \caption{a) Area, b) Power, and c) Accuracy comparison of our approximate printed MLPs vs the stohastic~\cite{Weller:2021:printed_stoch} \orange{and approximate~\cite{Armeniakos:DATE2022:axml}} ones.}
    \label{fig:stochastic}
\end{figure}

In Fig.~\ref{fig:stochastic} we compare our framework against the state-of-the-art stochastic printed MLPs~\cite{Weller:2021:printed_stoch} and \orange{approximate MLPs of~\cite{Armeniakos:DATE2022:axml}.}
In Fig.~\ref{fig:stochastic}, we included only the common MLPs examined in our work and~\cite{Weller:2021:printed_stoch}.
\orange{Since in~\cite{Armeniakos:DATE2022:axml} only a few datasets are considered, we followed the respective methodology to generate the approximate MLPs of the additional (i.e., not included in~\cite{Armeniakos:DATE2022:axml}) datasets}.
Since the stochastic MLPs mainly feature high accuracy degradation, for our MLPs we used the $5$\% accuracy loss threshold (see Fig.~\ref{fig:barplots}c).
\orange{Similarly, for~\cite{Armeniakos:DATE2022:axml}, we selected the approximate designs that feature the lowest area (and power) and up to $5$\% accuracy loss}.\label{commentR1C5}
% Note that accuracy loss is reported with respect to the exact baseline MLPs (Table~\ref{tab:baselines}).
% \cite{Weller:2021:printed_stoch} examined also the PD MLP but achieved only $0.22$ accuracy (while we achieve above $0.89$) and thus PD is not included in our comparison.
As shown, our approximate bespoke MLPs outperform~\cite{Weller:2021:printed_stoch} \orange{and~\cite{Armeniakos:DATE2022:axml}} at all the metrics examined (i.e., power, area, accuracy).
% Specifically, we achieve $3.5$x lower area, $4.2$x lower power, and $6.6$x lower accuracy loss than stochastic circuits~\cite{Weller:2021:printed_stoch}, on average.
Specifically, we achieve \yellow{$3.4$x} lower area, \yellow{$3.7$x} lower power, and \yellow{$7.7$x} lower accuracy loss than stochastic circuits~\cite{Weller:2021:printed_stoch}, on average.
Similarly, compared to~\cite{Armeniakos:DATE2022:axml}, our gains increase to \yellow{$8.8$x} lower area, \yellow{$7.8$x} lower power, and \yellow{$1.2$x} lower accuracy loss.
All the aforementioned gains refer to similar performance since \cite{Weller:2021:printed_stoch} required a stochastic bitstream of length $1024$.
In Fig.~\ref{fig:stochastic}, our MLPs require $200$ms per inference, while~\cite{Weller:2021:printed_stoch} requires $220$ms to $230$ms.
\orange{As in our case, the approximate designs of~\cite{Armeniakos:DATE2022:axml} operate at $200$ms per inference.}
\orange{Finally, we should mention that our co-design framework and~\cite{Weller:2021:printed_stoch} require some extra training epochs while~\cite{Armeniakos:DATE2022:axml} proposed a post-training approximation procedure.
Though, since printed electronics allow only relatively small MLPs with a 
limited number of parameters, the time overhead required for re-training is still negligible.
}

\section{Conclusion}\label{sec:conc}

Printed electronics offers a promising solutions to address limitations of silicon-based systems w.r.t. applications that require low cost, conformity, nontoxicity, etc.
However, the ultra-resource constraint nature of printed circuits prohibits the realization of complex circuits, such as machine learning classifiers.
In this work, we propose, for the first time, a software-hardware co-design framework for approximate printed MLPs.
Through our printed-friendly MLP retraining and approximate summation, we \label{commentR3C11}\yellow{design} for the first time high accuracy battery powered printed MLPs, paving the way towards smart complex printed applications.

\ifCLASSOPTIONcaptionsoff
  \newpage
\fi


% \end{thebibliography}

\bibliographystyle{IEEEtran}
\bibliography{references}

\end{document}